\documentclass[lettersize,journal]{IEEEtran}
\usepackage{amsmath,amsfonts,amssymb}
\usepackage{algorithmic}
\usepackage{array}
\usepackage{textcomp}
\usepackage{booktabs}
\usepackage{stfloats}
\usepackage{float}
\usepackage{cite}
\usepackage{verbatim}
\usepackage{graphicx}
\usepackage{multirow}
\usepackage{bbding}
\usepackage{hyperref}
\usepackage{xcolor}

\def\BibTeX{{\rm B\kern-.05em{\sc i\kern-.025em b}\kern-.08em
    T\kern-.1667em\lower.7ex\hbox{E}\kern-.125emX}}
\usepackage{balance}

\begin{document}
\title{MLC-GCN: Multi-Level Generated Connectome Based GCN for AD Analysis}
\author{Wenqi Zhu, Yinghua Fu, Ze Wang, for the Alzheimer's Disease Neuroimaging Initiative
	\thanks{Wenqi Zhu is with School of Optical-Electrical and Computer Engineering, University of Shanghai for Science and Technology, Shanghai, China (e-mail: 232280535@st.usst.edu.cn).}
	\thanks{Yinghua Fu is with Department of Diagnostic Radiology and Nuclear Medicine, University of Maryland School of Medicine, USA (e-mail: yinghua.fu@som.umaryland.edu).}
	\thanks{Ze Wang is with the Department of Diagnostic Radiology and Nuclear Medicine, University of Maryland School of Medicine, USA (e-mail: ze.wang@som.umaryland.edu).}
}
\markboth{Sample}
{Wenqi Zhu \MakeLowercase{\textit{et al.}}: MLC-GCN:  Multi-Level Generated Connectome Based GCN for AD Analysis}

\maketitle

\begin{abstract}
	\noindent Alzheimer's Disease (AD) is a currently incurable neurodegeneartive disease. Accurately detecting AD, especially in the early stage, represents a high research priority. AD is characterized by progressive cognitive impairments that are related to alterations in brain functional connectivity (FC). Based on this association, many studies have been published over the decades using FC and machine learning to differentiate AD from healthy aging. The most recent development in this detection method highlights the use of graph neural network (GNN) as the brain functionality analysis. In this paper, we proposed a stack of spatio-temporal feature extraction and graph generation based AD classification model using resting state fMRI. The proposed multi-level generated connectome (MLC) based graph convolutional network (GCN) (MLC-GCN) contains a multi-graph generation block and a GCN prediction block. The multi-graph generation block consists of a hierarchy of spatio-temporal feature extraction layers for extracting spatio-temporal rsfMRI features at different depths and building the corresponding connectomes. The GCN prediction block takes the learned multi-level connectomes to build and optimize GCNs at each level and concatenates the learned graphical features as the final predicting features for AD classification. Through independent cohort validations, MLC-GCN shows better performance for differentiating MCI, AD, and normal aging than state-of-art GCN and rsfMRI based AD classifiers. The proposed MLC-GCN also showed high explainability in terms of learning clinically reasonable connectome node and connectivity features from two independent datasets. While we only tested MLC-GCN on AD, the basic rsfMRI-based multi-level learned GCN based outcome prediction strategy is valid for other diseases or clinical outcomes.
	
\end{abstract}


\section{Introduction}
\label{sec:0}
\IEEEPARstart{A}{lzheimer's} disease (AD) is a progressive neurodegenerative disease characterized by hallmark pathological depositions and cognitive impairments such as memory decline and executive dysfunction \cite{alzheimer20172017}. Due to the lack of an effective cure for AD and the unclear etiology, a top research focus is on early disease diagnosis as the best hope of treatment or interventions for AD is to delay or slow down its progression in the early stage. Functional magnetic resonance imaging (fMRI) is a non-invasive imaging technique that has been increasingly used in AD research due to its ability to probe regional brain function alterations and interregional functional connectivity (FC) and subsequently the whole brain connectome \cite{dennis2014functional, lang2014resting}. Over the past decade, several studies have investigated resting state fMRI (rsfMRI)-revealed FC in AD diagnosis using traditional classification methods such as generalized linear regression \cite{teipel2017robust}, random forest \cite{koch2012diagnostic} and support vector machine (SVM) \cite{dyrba2015multimodal}. These methods are often limited by the dependence of prior knowledge, the empirical and complicated feature selection as well as the  inability to extract and use data features at different hierarchies, which leads to sub-optimal classification performance.
The most recent development in this research topic highlights the use of  deep neural network (DNN), which is the state-of-art in machine learning that is free from the above-mentioned drawbacks of traditional "shallow" machine learning. The initial work of DNN in AD prediction mainly is based on convolutional neural networks (CNNs), a popular and powerful DNN framework \cite{kam2018novel, ramzan2020deep, parmar2020spatiotemporal}. Ramzan et al. used ResNet-18 to extract image features from 2D fMRI image slices to build an AD multi-classification predictor \cite{ramzan2020deep}. Kam et al. built a 3D CNN AD classifier using the 3D fMRI images as the input \cite{kam2018novel}. Parmar et al. used the 4D rsfMRI data (five consecutive rsfMRI image volumes) to build a CNN-based multi-class (normal aging, early and late mild cognitive impairment, and AD) AD classifier \cite{parmar2020spatiotemporal}. FC was not explicitly considered in these studies although the ICA process included in \cite{kam2018novel} can implicitly utilize FC.

The whole brain FC matrix (connectome) has long been used to model the brain connection network in a graph defined by the spatial node and inter-node connections. A natural way to combine connectome and DNN for AD or other disease prediction is to build a network graph using the functional connectome and then input the graph to a graph convolutional network (GCN) \cite{parisot2017spectral,kazi2019inceptiongcn,song2020integrating,huang2020edge, xing2021ds, yao2021mutual, yang2023mapping}. In an early study \cite{parisot2017spectral}, Parisot et al. presented a group level GCN including individual rsfMRI data and phenotypic data to form a population level graph. Functional connectome is included as the node feature and subsequently condensed using graph Fourier transform in the populational graph. Kazi et al.\cite{kazi2019inceptiongcn} proposed an extension of the GCN by Parisot et al. \cite{parisot2017spectral} through changing the receptive field of the filters so that the inter- and intra-graph heterogeneity can be better captured. Song et al. \cite{song2020integrating} proposed an improved GCN by changing the node similarity calculation to consider the disease status difference between different categories and similarity difference between data modalities. Another variation of the GCN by Parisot et al was proposed by Huang and Chung \cite{huang2020edge}; they used Monte Carlo simulations to drop out non-effective edges in the population graph. Song et al. \cite{song2022multicenter} incorporated multi-center and structural and functional connectivity into the population graph. Instead of directly using the individual subject's FC as the input features of the population GCN, Jiang et al. \cite{jiang2020hi} used a pre-processing GCN to condense the FC features of each individual and then input them to the population GCN. Zhang et al. \cite{zhang2023classification} published a similar approach but used an attention module to select the top k nodes from the individual level GCN output as the input features to the population GCN. These combined individual and population GCN approach may be able to extract better represented individual FC features and reduce the dimension of the input data to the population GCN but they should still be grouped into the population GCNs. Overall, the population GCNs achieve encouraging AD and other disease prediction accuracy, which may be a significant contribution by the inter-subject relationship encoding through the phenotypic data enhanced inter-node (inter-subject) association and populational graph learning. A big limitation of these methods is that the population graph needs to be created from the entire cohort which limits the generalizability and scalability. Pooling all subjects together and learning their features for classification in a single network is similar to a global data decomposition process, which risks being overfitted to the included subjects. The explicit inter-subject relationship encoding may further escalate this issue. When the number of subjects increases, there is an exponential increase of computation complexity. Meanwhile, the populational graph would need to be regenerated for a new subject and need to be retrained, which is nearly impractical. Another issue is that the population GCNs are difficult to interpret in the original brain space. These issues can be addressed using the individual GCNs.

Recently, a few individual GCN studies have been published \cite{xing2021ds, yao2021mutual, yang2023mapping}. In Xing et al. \cite{xing2021ds}, the dynamic brain connectome-generated GCNs were used for AD and biological value prediction. Yao et al. \cite{yao2021mutual} then proposed a mutual learning-based multi-scale triplet-based GCN to combine structural connectome and functional connectome for various brain disorder prediction. Yang et al \cite{yang2023mapping} published a similar but simpler functional connectome and structural connectome and mutual learning based GCN based brain disorder classifier. Instead of using the Pearson correlation as the connectome association to form the network structure, a few studies have proposed network structure learning strategies through modeling the potential nonlinear spatio-temporal inter-regional relationship \cite{zhang2019new, kan2022fbnetgen, yu2022learning}. Thus far, only the lowest level connectivity has been considered.

Neuronal signal is known to have a multi-scale structure, as does inter-regional connectivity. To investigate the multi-scale multi-level property of functional connectome, we propose a novel multi-level feature extraction based GCN working on individual subject's rsfMRI data. We dub this technique as the multi-level generated connectome GCN (MLC-GCN). To generate the connectomes at different scales of the input BOLD signals, we introduce a hierarchy of spatio-temporal feature extractors (STFEs) to extract spatio-temporal representations of the input BOLD signals at different level and use them to generate connectomes at different levels. The generated connectomes are then input into multiple GCNs to further learn the feature representation, and outputs of all different GCNs are concatenated and finally sent to a classifier to predict the disease status. The multi-classification task involves more discriminative features than bi-classification, so we take AD multi-classification to train the proposed MLC-GCN architecture. The contributions of this paper are summarized as follows:

\begin{enumerate}
	\item{A novel GCN architecture (MLC-GCN) for AD multi-classification is proposed to obtain rich temporal and regional correlations by combining the generated conectomes at different levels and multiple independent GCN encoders.}
	\item{STFE is specially designed to extract spatio-temporal features for different levels of time series data, which includes more significant multi-scale information than previous architectures.}
	\item{The experimental results on the public medical datasets Alzheimer's Disease Neuroimaging Initiative (ADNI) and Open Access Series of Imaging Studies-3 (OASIS-3) demonstrate that MLC-GCN achieves the state-of-the-art performance. Extensive ablation experiments are also conducted to discuss the effectiveness of modules in MLC-GCN.}
\end{enumerate}

The remainder of this paper is organized as follows. Section \ref{sec:2} presents the architecture of MLC-GCN in detail. Section \ref{sec:3} describes the experimental results of MLC-GCN on the datasets ADNI and OASIS-3 compared with other methods, the ablation experiments are described to indicate the effectiveness of the proposed modules, and the generated graphs in the multi-classification task are visualized and analyzed for association with existing brain research related to AD. Finally, we conclude the paper and offer suggestions for further study in Section \ref{sec:5}.

\section{Methods}
\label{sec:2}
\begin{figure*}[!t]
	\centering
	\includegraphics[width=2\columnwidth]{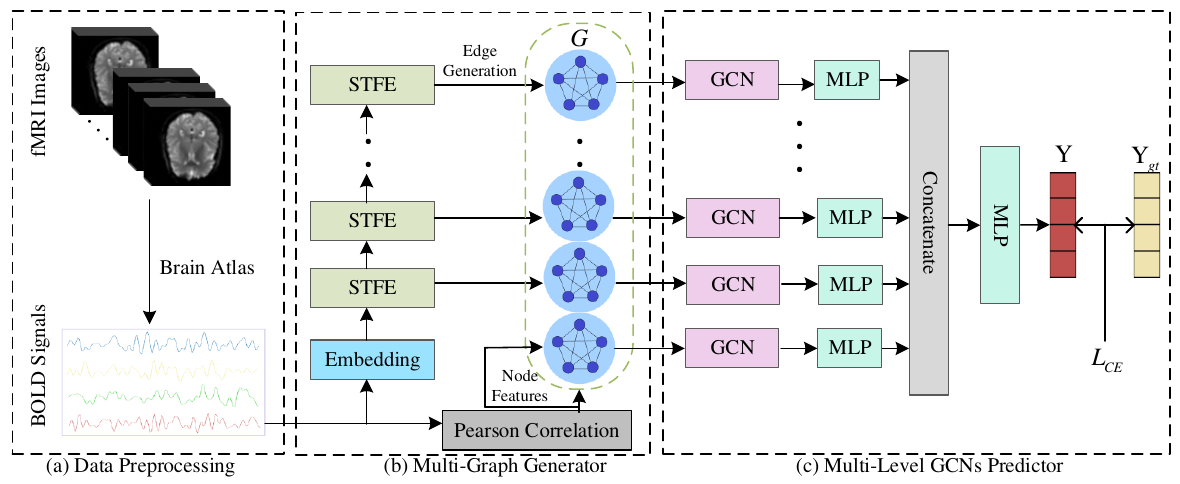}
	\caption{The overall workflow of the MLC-GCN contains: a data preprocessing module (a), a multi-graph generator (b), and a multi-level GCN-based predictor. In module a, a brain atlas is used to extract n time series from n brain regions. In b, temporal embedding and STFE are used to extract sparse temporal features at different hierarchy and to calculate the corresponding graphs (connectomes). In c, GCNs are used to encode the generated graphs at different levels into higher level graphical data features. These features are concatenated and input to a multi-layer perceptron (MLP) to classify AD.}
	\label{fig.1}
\end{figure*}

Fig. \ref{fig.1} illustrates the flowchart of  the proposed MLC-GCN. The first module (a) is for fMRI preprocessing. The right two modules are the multi-graph generator and the multi-level GCNs predictor included in the MLC-GCN. The multi-graph generator is designed to construct a brain connectome using the fMRI time series extracted through the embedding and STFE module at each different level. Each generated connectome (graph) is then encoded into an embedding vectors by an independent GCN. The output of all GCNs are concatenated into a vector which is input to a multi-layer perceptron (MLP) for predicting AD status (or other clinical outcome for other disease).

\subsection{Multi-Graph Generator}
\label{sec:2.1}
The multi-graph generator contains a hierarchy of feature extraction and graph generation module. At the lowest level, the preprocessed fMRI time series with a length of $L$ from n ROIs are directly used to calculate a $n\times n$ Pearson correlation coefficient matrix. Matrix element at the $i-$th row and $j-$th column is the correlation coefficient between the fMRI time series of the $i-$th ROI and the $j-$th ROI ($i$, and $j$ are from 1 to $n$). This matrix is used to build the brain graph: the connectome $F$. At each higher level of the hierarchy $i$, the output of the lower level STFE module is sent to a new STFE module to extract new temporal features of the fMRI signal $h_i$. These features will be used to calculate a correlation coefficient matrix and form a graph (the brain connectome) $A^{(i)}$. The temporal features are also sent to the upper level for further processing.  At the second level, an embedding layer is added in front of the STFE module, which is made of a 1D-CNN designed to extract the compact feature representation for each of all $n$ time series in the successive STFE module and to form the feature matrix with size $n\times l$  ($l$ $\leq L$).  Details are provided below.

\subsubsection{Input Embedding}
\label{sec:2.1.1}
The embedding layer is used to extract abstract temporal features from the discretely sampled points. For ROI $p$, denote the corresponding time series by $X_p$ (length is $L$). After 1D temporal convolution through multiple kernels in the 1D-CNN, a new series $Z_p$ with a length of $l$ is obtained. For the preprocessed BOLD signals from fMRI $X = \{ {X_1}, {X_2}, \cdots, {X_n}\}$ where $n$ denotes the number of ROIs, the embedded feature of the whole brain $Z$ is obtained through a fully-connected layer with the following formulas:

\begin{equation}\label{eq.1}
	Z = \sigma (Flatten(Conv(X))W) + PE
\end{equation}

\begin{equation}\label{eq.2}
	\begin{aligned}
		PE(pos,2q) &= \sin (\frac{{pos}}{{{{10000}^{2q/h}}}}), \\
		PE(pos,2q + 1) &= \cos (\frac{{pos}}{{{{10000}^{2q/h}}}}) \\
	\end{aligned}
\end{equation}
where $Conv()$ denotes a 1D-CNN with size $t$ and $m$ kernels involving, $W \in {R^{mL \times l}}$ the learnable weight matrix of the linear layer, $\sigma (\cdot)$ the activation function, $Z=\{ {Z_1}, {Z_2}, \cdots, {Z_n}\}$ the hidden representation of time series. ${PE}$ in \eqref{eq.2} is the constant position embedding \cite{2017Attention}, where $pos$ is the token position and $q$ is the embedding dimension which is set to be $n$ in this paper. $X\in{R^{n \times L}}$ is transformed into the hidden representation $Z\in{R^{n \times l}}$, where ${Z_p}$ is the embedded vector of the $p-$ ROI.

\subsubsection{The STFE Module}
\label{sec:2.1.2}

\begin{figure}[!t]
	\centering
	\includegraphics[width=1.0\columnwidth]{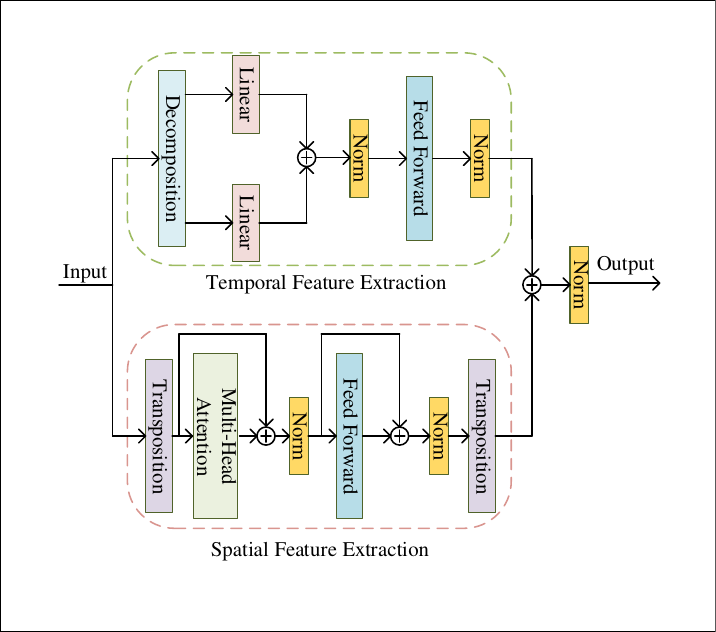}
	\caption{The framework of the proposed STFE module.}
	\label{fig.2}
\end{figure}

Fig. \ref{fig.2} illustrates the architecture of a STFE module, which consists of two parallel pathways: one for spatial feature extraction (SFE), the other for temporal feature extraction (TFE). The SFE pathway is composed of the encoder of a transformer \cite{2017Attention}, consisting of a multi-head attention layer and a feed forward layer. Given the input feature $h_{i-1}$ which is the output of the preceding level STFE, SFE at the $i-$th level can be described as:

\begin{equation}\label{eq.4}
	{h_{Trans}} = Trs({h_{i-1}^T})
\end{equation}
where $Trs()$ means the encoder of a transformer consisting of a multi-head attention layer and a multilayer perceptron (MLP) layer. We use a transformer in the SFE pathway to specifically encode the node order information during spatial feature extraction.

The TFE pathway contains a linear decomposition block and a feed forward layer connected and followed by a normalization (Norm) block. The linear decomposer (DLinear) is designed to extract the hidden features through a trend-cycle operator and seasonal variation operator \cite{zeng2023transformers}, which has shown good performance for feature extraction for univariate time series. For the input feature $h_{i-1}$, TFE will perform the following operations:

\begin{equation}\label{eq.3}
	\begin{aligned}
		{h_{trend}} &= AvgPool(h_{i-1}), \\
		{h_{seasonal}} &= h_{i-1} - {h_{trend}}, \\
		{h_{DLinear}} &= MLP(\sigma ({W_t}{h_{trend}}) + \sigma ({W_s}{h_{seasonal}})) \\
	\end{aligned}
\end{equation}
where ${h_{trend}}$ refers to the trend-cycle features, ${h_{seasonal}}$ means the seasonal variation features, ${W_t}$ and ${W_s}$ are the learnable matrices. $AvgPool()$ with size $t$ is implemented with padding operation to keep the size of $h_{i-1}$.

The output of SFE and TFE are then fused to form the output of the $i-$th level STFE and can be described through Equations \eqref{eq.5} and \eqref{eq.6} below:

\begin{equation}\label{eq.5}
	h_i = MLP({h_{DLinear} + h_{Trans}}^T)
\end{equation}

\begin{equation}\label{eq.6}
	h_{i+1} = STFE(h_i), h_0 = Z \\
\end{equation}
where $h_i$, $i=0, 1, 2, \cdots, K$, denotes the output of the $i-$th layer, and $k$ is the total number of STFE blocks.

After normalization, ${h^i}$ can be used to build the $i-$th level connectome (the adjacency matrix $A$) through dot product:

\begin{equation}\label{eq.6-1}
	A^{(i)} = h_i(h_i)^T
\end{equation}

Through the $K$ levels of STFE, we will get $K$ sets of $n$ features (note that we keep the spatial dimension after feature extraction through the SFE pathway): $h=\{ h_1, h_2, \cdots, h_K\}$ and the corresponding set of  $K$ adjacency matrices: $G=\left\{ {{G_1}, {G_2}, \cdots, {G_K}}\right\}$.

\subsection{Multi-Level GCNs-based Predictor}
\label{sec:2.2}
The standard brain connectome $G_0$ (calculated from the preprocessed fMRI time series) and the generated graphs $G=\left\{ {{G_1}, {G_2}, \cdots, {G_K}} \right\}$ at $K$ levels are sent to a $K+1$ level GCNs-based predictor as the input. For each of the $K+1$ levels, a GCN will be learned based on the corresponding input ${G_i}$, $i=0, 1, 2, \cdots, K$  to generate the output graph embedding ${E_i}$, $i=0, 1, 2, \cdots, K$. Denote the input temporal features of the nodes for the $j-$th layer of the  $i-$th GCN by $h^{(i)}_j$: $h^{(i)}_j= [h^{(i)}_{j1}, h^{(i)}_{j2}, \cdots, h^{(i)}_{jn}]^T$. The operation of the $j-$th layer of this GCN can be described by:

\begin{equation}\label{eq.8}
	\begin{aligned}
		h^{(i)}_{j + 1} &= \sigma (\widehat{A^{(i)}}h^{(i)}_jW^{(i)}_j), \\
		h^{(i)}_0 &= F, \widehat{A^{(i)}} = A^{(i)} + I \\
	\end{aligned}
\end{equation}
where $A^{(i)} \in {R^{n \times n}}$ denotes the adjacency matrix of the generated graph, $n$ is the number of nodes, $I$ is an identity matrix performing as self-connections, $W^{(i)}_j$ is a trainable weight matrix of the $j$th layer and $\sigma ( \cdot )$ the activation function. $F$ is Pearson correlation coefficient matrix. For simplicity, the same adjacency matrix $A^{(i)}$ is used for all layers of GCNs, and the number of layer of GCN is set to 2.

The output of the last graph convolutional layer of the $k-$th GCN will be embedded into a 1D vector $E_k$ through an MLP operation and concatenated according to the order of $k$ in all $K$ GCNs into a the final multi-level graph embedded vector $E = \left\{ {{E_0}, {E_1}, \cdots, {E_k}} \right\}$. This fused vector is then passed to a MLP for clinical outcome prediction:

\begin{equation}\label{eq.9}
	Y = Softmax(MLP(Concat({E_0}, {E_1}, \cdots, {E_k})))
\end{equation}
where $Y = [{y_0}, {y_1}, \cdots, {y_c}]^{T}$ is the final output of MLC-GCN referring to the disease state of the input images, $c$ the number of class categories.

\subsection{Objective Function}
\label{sec:2.3}
For prediction performance, we adopt the cross entropy loss $L_{CE}$ to constraint MLC-GCN:

\begin{equation}\label{eq.10}
	{L_{CE}} =  - \frac{1}{N}\sum\limits_{i = 1}^c {1(y = i)\log (\hat y)}
\end{equation}
where $\hat y$ and $y$ represent the predicted results and the ground-truth labels, respectively. $N$ is the number of samples and $1(\cdot)$ is the indicator function.

To force the graph generator of the MLC-GCN to learn mutually different connectomes and the corresponding graphs, we add an intra-group loss to minimize the intra-group difference as described by \eqref{eq.7} below:

\begin{equation}\label{eq.7}
	\begin{aligned}
		{L_{group}} &= \frac{1}{K}\sum\limits_{i = 1}^k {\sum\limits_{c \in C} {\sum\limits_{u \in {S^c}} {\frac{{\left\| {A_u^{(i)} - \mu _c^{(i)}} \right\|_2^2}}{{\left| {{S^c}} \right|}}} } } , \\
		\mu _c^{(i)} &= \sum\limits_{u \in {S^c}} {\frac{{A_u^{(i)}}}{{\left| {{S^c}} \right|}}}  \\
	\end{aligned}
\end{equation}
where $C$ denotes the set of class labels; ${S^c} = \{ u|{Y_{u,c}} = 1\}$ is the set of samples with label $c$; $K$ is the number of levels of STFE; ${\mu _c^{(i)}}$ is the mean of the learnable adjacency matrix ${A_u^{(i)}}$ in class $c$ and layer $i$ within a batch.

The final loss is a combination of $L_{CE}$ and the intra-group loss $L_{group}$:

\begin{equation}\label{eq.11}
	{L_{total}} = {L_{CE}} + \alpha {L_{group}}
\end{equation}
where $\alpha$ is hyper-parameter and is set to 1.0 in this paper.

\section{Experiments, Results and Discussion}
\label{sec:3}
In this section, we present several comparative experiments and an ablation study on the public datasets ADNI and OASIS-3 to validate the effectiveness of MLC-GCN. Furthermore, we investigate the brain graphs generated by MLC-GCN to explore their consistency with existing neuroscience discoveries.

\subsection{Datasets}
\label{sec:3.1}
The demographic statistics of  ADNI and OASIS-3  are illustrated in Table. \ref{tab.1}

\begin{table}[H]
	\begin{center}
		\caption{The demographic statistics of the datasets used in this work.}
		\centering
		\label{tab.1}
		\resizebox{\linewidth}{!}{
			\begin{tabular}{c|ccccc}
				\toprule
				\hline
				Dataset & Label & Scans Number & Patients number & Gender (M/F) & Age (mean$\pm$std.) \\
				\hline
				\multirow{4}{*}{ADNI} & NC & 187 & 52 & 30/22 & 75.01$\pm$6.23 \\
				& EMCI & 182 & 55 & 28/27 & 72.56$\pm$6.00 \\
				& LMCI & 156 & 40 & 13/27 & 72.28$\pm$7.60 \\
				& AD & 118 & 34 & 16/18 & 74.69$\pm$7.47 \\
				\hline
				\multirow{4}{*}{OASIS-3} & NC & 570 & 225 & 85/140 & 68.24$\pm$8.94 \\
				& MCI & 225 & 96 & 58/38 & 76.90$\pm$7.49 \\
				& AD & 105 & 50 & 27/23 & 76.83$\pm$8.03 \\
				\hline
				\bottomrule
			\end{tabular}
		}
	\end{center}	
\end{table}

\subsubsection{ADNI}
Alzheimer's Disease Neuroimaging Initiative (ADNI) is a large-scale dataset and contains longitudinal brain MRI for AD study. We collected a sub-dataset with 643 fMRI which were acquired with a 3T MRI scanner (Philips Medical Systems,  Cambridge, MA) with specific scanning parameters containing a TR/TE of 3000 ms/30 ms, imaging matrix of 64 $\times$ 64, voxel size of 3.3 mm $\times$ 3.3 mm $\times$ 3.3 mm, and 48 slices. The chosen data includes 187 NCs, 118 ADs, 182 early mild cognitive impairments (EMCIs) and 156 late mild cognitive impairments (LMCIs). In the multi-classification experiment, a four classification task will be performed.

\subsubsection{OASIS-3}
Open Access Series of Imaging Studies-3 (OASIS-3) is a compilation of MRI and PET imaging and related clinical data collected across several ongoing studies in the Washington University Knight Alzheimer Disease Research Center over the course of 15 years \cite{lamontagne2019oasis}. The acquisition was performed using a 3T MRI scanner  manufactured by Siemens with specific scanning parameters containing a TR/TE of 2200 ms/27 ms, imaging matrix of 64 $\times$ 64, voxel size of 4.0 mm $\times$ 4.0 mm $\times$ 4.0 mm, and 36 slices. We selected 900 samples of fMRI and labeled each  datum based on the clinical measurement rate (CDR) \cite{morris1993clinical}, including 570 with $CDR=0$, 225 with $CDR=0.5$ and 105 with $CDR>0.5$, which are respectively considered as NC, MCI and AD labels. All three categories are used for the multi-classification task.

\begin{table*}[!t]
	\begin{center}
		\caption{Binary classification of different methods on the ADNI (NC vs AD)}
		\centering
		\label{tab.2}
		\begin{tabular*}{0.7\linewidth}{c|ccccc}
			\toprule
			\hline
			Method & Acc & AUC & Spe & Sen & F1-score \\
			\hline
			Random Forest \cite{koch2012diagnostic} & 83.28$\pm$5.60 & 80.15$\pm$5.61 & 95.69$\pm$1.50 & 73.49$\pm$3.61 & 74.60$\pm$4.06 \\
			SVM \cite{dyrba2015multimodal} & 87.21$\pm$3.15 & 86.51$\pm$1.06 & 87.65$\pm$0.57 & 86.11$\pm$3.47 & 95.93$\pm$3.29 \\
			\hline
			DNN \cite{heinsfeld2018identification} & 92.46$\pm$2.74 & 94.03$\pm$2.49 & 96.36$\pm$3.22 & 91.27$\pm$3.71 & 91.84$\pm$3.16 \\
			BrainnetCNN \cite{kawahara2017brainnetcnn} & 91.80$\pm$3.06 & 93.29$\pm$5.54 & 93.02$\pm$4.13 & 89.33$\pm$3.23 & 89.55$\pm$3.40 \\
			FCNet \cite{riaz2017fcnet} & 90.16$\pm$3.48 & 91.70$\pm$6.33 & 93.60$\pm$3.58 & 89.17$\pm$3.87 & 89.51$\pm$3.74 \\
			\hline
			GAT \cite{velivckovic2018graph} & 89.18$\pm$2.97 & 91.92$\pm$2.57 & 91.41$\pm$3.59 & 88.53$\pm$2.90 & 88.59$\pm$3.03 \\
			GCN \cite{kipf2017semi} & 91.47$\pm$3.55 & 92.62$\pm$1.39 & 97.34$\pm$3.75 & 89.75$\pm$3.98 & 90.72$\pm$3.94 \\
			BrainGNN \cite{li2021braingnn} & 87.86$\pm$1.87 & 91.31$\pm$2.65 & 91.41$\pm$2.19 & 86.86$\pm$1.93 & 87.15$\pm$2.03 \\
			MMTGCN \cite{yao2021mutual} & 90.38$\pm$2.65 & 91.48$\pm$1.78 & 95.55$\pm$3.73 & 86.01$\pm$1.37 & 87.05$\pm$3.54 \\
			FBNetGNN-GRU \cite{kan2022fbnetgen} & 93.11$\pm$3.56 & 94.45$\pm$4.09 & 96.76$\pm$3.53 & 92.21$\pm$3.41 & 91.93$\pm$3.57 \\
			FBNetGNN-CNN \cite{kan2022fbnetgen} & 92.80$\pm$2.59 & 94.55$\pm$3.17 & 95.84$\pm$2.27 & 91.43$\pm$3.19 & 91.54$\pm$2.74 \\
			DABNet \cite{yu2023deep} & 93.44$\pm$2.84 & 95.06$\pm$3.44 & 98.36$\pm$2.43 & 91.59$\pm$3.61 & 92.81$\pm$3.14 \\
			LG-GNN \cite{zhang2023classification} & 93.44$\pm$2.53 & 95.17$\pm$3.35 & 97.85$\pm$2.02 & 92.48$\pm$3.60 & 91.06$\pm$3.46 \\
			\hline
			${\text{MLC-GC}}{{\text{N}}_6}$ & 94.10$\pm$2.49 & 95.36$\pm$3.27 & 98.47$\pm$1.18 & 92.53$\pm$3.17 & 93.56$\pm$2.75 \\
			${\text{MLC-GC}}{{\text{N}}_{12}}$ & 95.08$\pm$1.16 & 96.25$\pm$3.27 & \textbf{98.99$\pm$1.38} & 94.05$\pm$1.40 & 94.71$\pm$1.26 \\
			${\text{MLC-GC}}{{\text{N}}_{24}}$ & \textbf{95.74$\pm$0.90} & \textbf{97.76$\pm$2.17} & 98.98$\pm$1.40 & \textbf{95.08$\pm$1.17} & \textbf{95.46$\pm$.096} \\
			\hline
			\bottomrule
		\end{tabular*}
	\end{center}	
\end{table*}

\begin{table*}[!t]
	\begin{center}
		\caption{Multi-classification of different methods on ADNI and OASIS-3}
		\centering
		\label{tab.3}
		\begin{tabular*}{0.8\linewidth}{c|c|ccccc}
			\toprule
			\hline
			Dataset & Method & Acc & AUC & Spe & Sen & F1-score \\
			\hline
			\multirow{12}{*}{ADNI} & Random Forest \cite{koch2012diagnostic} & 54.42$\pm$3.69 & 80.29$\pm$2.90 & 84.20$\pm$1.33 & 51.28$\pm$3.66 & 51.13$\pm$3.89 \\
			& SVM \cite{dyrba2015multimodal} & 63.39$\pm$1.65 & 86.51$\pm$1.06 & 87.65$\pm$0.57 & 63.75$\pm$1.53 & 63.36$\pm$1.67 \\
			& DNN \cite{heinsfeld2018identification} & 72.27$\pm$2.43 & 88.31$\pm$2.83 & 90.76$\pm$0.78 & 72.26$\pm$1.95 & 72.99$\pm$2.10 \\
			& BrainnetCNN \cite{kawahara2017brainnetcnn} & 72.51$\pm$5.03 & 88.18$\pm$4.18 & 90.65$\pm$1.68 & 72.61$\pm$4.77 & 72.87$\pm$4.98 \\
			& FCNet \cite{riaz2017fcnet} & 69.20$\pm$3.82 & 86.27$\pm$1.50 & 89.56$\pm$1.26 & 69.31$\pm$3.85 & 69.24$\pm$4.09 \\
			& GAT \cite{velivckovic2018graph} & 67.04$\pm$4.09 & 83.56$\pm$3.16 & 88.58$\pm$1.59 & 66.41$\pm$4.65 & 66.53$\pm$4.62 \\
			& GCN \cite{kipf2017semi} & 73.56$\pm$2.63 & 88.68$\pm$1.69 & 91.07$\pm$0.91 & 73.71$\pm$2.62 & 73.67$\pm$2.51 \\
			& BrainGNN \cite{li2021braingnn} & 66.83$\pm$3.45 & 84.49$\pm$3.72 & 89.17$\pm$1.73 & 66.74$\pm$4.07 & 66.54$\pm$3.85 \\
			& FBNetGNN-GRU \cite{kan2022fbnetgen} & 73.09$\pm$2.64 & 88.76$\pm$2.44 & 88.86$\pm$4.47 & 73.02$\pm$3.55 & 73.09$\pm$2.99 \\
			& FBNetGNN-CNN \cite{kan2022fbnetgen} & 71.85$\pm$4.31 & 88.00$\pm$2.72 & 90.26$\pm$1.09 & 71.89$\pm$4.44 & 72.12$\pm$4.38 \\
			& LG-GNN \cite{zhang2023classification} & 76.19$\pm$3.87 & 90.84$\pm$4.52 & 91.87$\pm$1.29 & 75.47$\pm$4.52 & 76.01$\pm$4.29 \\
			& ${\text{MLC-GC}}{{\text{N}}_6}$ & 79.03$\pm$5.11 & 92.33$\pm$2.84 & 92.88$\pm$1.71 & 79.21$\pm$4.79 & 79.41$\pm$5.14 \\
			& ${\text{MLC-GC}}{{\text{N}}_{12}}$ & 80.24$\pm$4.78 & 92.23$\pm$2.41 & 93.28$\pm$1.65 & 80.38$\pm$5.15 & 80.57$\pm$4.73 \\
			& ${\text{MLC-GC}}{{\text{N}}_{24}}$ & \textbf{82.26$\pm$4.61} & \textbf{92.48$\pm$2.72} & \textbf{95.16$\pm$2.14} & \textbf{82.38$\pm$4.65} & \textbf{82.51$\pm$4.46} \\
			\hline
			\multirow{12}{*}{OASIS-3} & Random Forest \cite{koch2012diagnostic} & 68.22$\pm$0.91 & 80.13$\pm$3.58 & 72.72$\pm$0.46 & 41.84$\pm$1.54 & 41.14$\pm$2.85 \\
			& SVM \cite{dyrba2015multimodal} & 76.44$\pm$2.80 & 85.91$\pm$2.79 & 81.99$\pm$1.32 & 55.82$\pm$4.09 & 57.89$\pm$6.24 \\
			& DNN \cite{heinsfeld2018identification} & 84.22$\pm$2.24 & 89.20$\pm$1.78 & 88.59$\pm$1.46 & 76.19$\pm$2.91 & 77.93$\pm$4.27 \\
			& BrainnetCNN \cite{kawahara2017brainnetcnn} & 85.56$\pm$2.12 & 91.16$\pm$1.81 & 88.69$\pm$2.01 & 75.21$\pm$3.32 & 79.19$\pm$2.25 \\
			& FCNet \cite{riaz2017fcnet} & 78.44$\pm$4.02 & 85.87$\pm$3.98 & 86.30$\pm$3.09 & 70.22$\pm$6.61 & 70.12$\pm$5.92 \\
			& GAT \cite{velivckovic2018graph} & 79.33$\pm$1.44 & 85.62$\pm$2.38 & 85.42$\pm$1.27 & 67.94$\pm$1.82 & 70.91$\pm$1.77 \\
			& GCN \cite{kipf2017semi} & 83.33$\pm$0.88 & 89.68$\pm$1.14 & 87.98$\pm$0.95 & 73.49$\pm$3.52 & 75.27$\pm$2.92 \\
			& BrainGNN \cite{li2021braingnn} & 76.98$\pm$2.41 & 85.16$\pm$3.42 & 84.80$\pm$2.22 & 67.77$\pm$3.52 & 67.84$\pm$2.98 \\
			& FBNetGNN-GRU \cite{kan2022fbnetgen} & 85.89$\pm$2.44 & 90.37$\pm$1.73 & 90.14$\pm$1.81 & 77.68$\pm$5.16 & 79.92$\pm$4.04 \\
			& FBNetGNN-CNN \cite{kan2022fbnetgen} & 85.67$\pm$2.87 & 90.58$\pm$2.31 & 89.70$\pm$1.37 & 77.96$\pm$4.82 & 79.19$\pm$5.44 \\
			& LG-GNN \cite{zhang2023classification} & 89.33$\pm$1.58 & 94.05$\pm$1.67 & 92.04$\pm$0.79 & 81.90$\pm$1.81 & 84.41$\pm$2.59 \\
			& ${\text{MLC-GC}}{{\text{N}}_6}$ & 88.44$\pm$2.76 & 93.23$\pm$1.38 & 92.06$\pm$1.78 & 82.60$\pm$3.31 & 84.06$\pm$3.60 \\
			& ${\text{MLC-GC}}{{\text{N}}_{12}}$ & 90.11$\pm$1.38 & 94.10$\pm$0.86 & 92.70$\pm$1.11 & \textbf{83.46$\pm$3.04} & \textbf{86.01$\pm$2.85} \\
			& ${\text{MLC-GC}}{{\text{N}}_{24}}$ & \textbf{90.56$\pm$1.30} & \textbf{94.36$\pm$1.24} & \textbf{93.07$\pm$0.45} & 82.80$\pm$2.63 & 85.78$\pm$2.76 \\
			\hline
			\bottomrule
		\end{tabular*}
	\end{center}	
\end{table*}

\begin{table*}[!t]
	\begin{center}
		\caption{Multi-classification results of different ablated modules on ADNI.}
		\centering
		\label{tab.4}
		\begin{tabular*}{0.8\linewidth}{c|ccc|ccccc}
			\toprule
			\hline
			\multirow{2}{*}{Method} & \multicolumn{3}{c|}{Modules} & \multirow{2}{*}{Acc} & \multirow{2}{*}{AUC} & \multirow{2}{*}{Spe} & \multirow{2}{*}{Sen} & \multirow{2}{*}{F1-score} \\
			\cline{2-4}
			& TFE & SFE & ${L_{group}}$ & & & & & \\
			\hline
			\multirow{6}{*}{MLC-GCN} & & \checkmark & \checkmark & 78.22$\pm$5.08 & 92.08$\pm$2.59 & 92.59$\pm$1.72 & 78.49$\pm$5.43 & 78.68$\pm$5.36 \\
			& \checkmark & & \checkmark & 76.66$\pm$4.93 & 91.88$\pm$2.74 & 92.06$\pm$1.66 & 77.05$\pm$4.31 & 77.16$\pm$4.63 \\
			& \checkmark & \checkmark & & 78.53$\pm$4.66 & 91.93$\pm$2.87 & 92.76$\pm$1.74 & 79.11$\pm$5.28 & 78.99$\pm$4.93 \\
			& & \checkmark & & 77.91$\pm$5.01 & 91.98$\pm$2.64 & 92.44$\pm$1.73 & 78.27$\pm$5.42 & 78.58$\pm$5.10 \\
			& \checkmark & & & 76.87$\pm$4.54 & 91.65$\pm$2.76 & 91.89$\pm$1.88 & 77.19$\pm$5.02 & 77.14$\pm$4.43 \\
			& \checkmark & \checkmark & \checkmark & \textbf{79.03$\pm$5.11} & \textbf{92.33$\pm$2.84} & \textbf{92.88$\pm$1.71} & \textbf{79.21$\pm$4.79} & \textbf{79.41$\pm$5.14} \\
			
			\hline
			\bottomrule
		\end{tabular*}
	\end{center}	
\end{table*}

\subsection{Experiment Setting}
\label{sec:3.3}
MLC-GCN is implemented in PyTorch \cite{paszke2019pytorch} using a single NVIDIA RTX 4090 GPU. AdamW \cite{2018Decoupled} is chosen as the gradient descent optimization for training to automatically adjust the learning rate and update the variable by using the moving average value of the exponentially reduced gradient. The data augmentation method Mixup \cite{zhang2018mixup} is used to  enhance the generalization performance of the model. The weight decay is set to 0.001, the learning rate 0.001, the dropout rate 0.2, the maximum number of epochs in all experiments 300, the kernel size $t$ 5 and the hidden size $h$ 64. The 5-fold stratified cross-validation and five metrics are taken to evaluate the classification performance including accuracy (Acc), area under the curve (AUC), specificity (Spe), sensitivity (Sen) and F1-score.

\subsection{Classification Results}
\label{sec:3.4}

All data are preprocessed using the Brainnetome Toolkit \cite{xu2018brant} in a general procedure with slice timing correction, realignment to the first volume, spatial normalization to Montreal Neurological Institute (MNI) space, regression of nuisance signals and temporal bandpass filtering (0.01-0.08 Hz). The Brainnetome Atlas \cite{fan2016human} is applied to divide the brain into 273 ROIs, and the time series of each ROI in fMRI is obtained by averaging all voxels in the region at each time point.

MLC-GCN is compared with  traditional machine learning methods, DNNs and GCNs. Traditional machine learning methods include Random Forest  with 1000 estimators \cite{koch2012diagnostic} and SVM with the RBF kernel \cite{dyrba2015multimodal}. DNNs include a dual-encoders CNN \cite{heinsfeld2018identification}, BrainnetCNN \cite{kawahara2017brainnetcnn} and FCNet \cite{riaz2017fcnet}. GCNs come from three different types: 4 individual level GCNs \cite{kipf2017semi}: GAT \cite{velivckovic2018graph}, BrainGNN \cite{li2021braingnn} and MMTGCN \cite{yao2021mutual}, a hybrid GNN LG-GNN \cite{zhang2023classification} as well as 2 generated GCNs: FBNetGNN \cite{kan2022fbnetgen} and DABNet \cite{yu2023deep}. Those networks were built using the open-source codes available in Github. The numerical values are presented as the form of (\%, mean$\pm$standard deviation) in all experiments. 

Table \ref{tab.2} presents the results of the bi-classification task differentiating NC from AD on ADNI. The  proposed  MLC-GCN in this paper was trained using 6, 12, and 24 levels of STFE denoting  MLC\_GCN$_6$,  MLC\_GCN$_{12}$ and  MLC\_GCN$_{24}$ respectively. It can be seen that the machine learning methods perform poorly compared to other types of methods due to the complexity of fMRI. CNN \cite{heinsfeld2018identification} has better performance in the classification between NC and AD than brain GNNs because the Pearson graph may include the unrelated brain information of AD, which even outperforms the SOTA brain GNN model MMTGCN \cite{yao2021mutual}.  LG-GCN \cite{zhang2023classification} as a hybrid GNN as well as FBNetGNN \cite{kan2022fbnetgen} and DABNet \cite{yu2023deep} as the graph generation method have got the better performance than brain GNNs and DNNs without graph. LG-GCN gets a little higher performance than FBNetGNN  in Acc, AUC, Spe and Sen, but has the similar performace with DABNet. Moreover, MLC-GCN with the basic 6-layer STFE is superior to all other methods on all metrics, and the performance also increases with the number of layers of STFE.

Table \ref{tab.3} gives the results of the multi-classification task on the ADNI and OASIS-3. In this case, the accuracy of machine learning methods is greatly reduced indicating that hand-crafted extractors is hard to capture the representative features when samples become increasingly complex. The performance of DNNs and CNNs also decreases with the highest accuracy lower than 75\% and FCNet even lower than 70\% on ADNI. Several GNN-based methods still outperform BrainnetCNN and DNN, which indicates that their architectures capture the representative features even with the increasing data complexity. It should be noted that FBNetGNN performs worse than GCN taking Pearson graph on ADNI, as the stronger feature extractors to construct the brain graph are needed when the data becomes more various and complex. LG-GCN still obtains the best performance among the published methods because of the sophisticated architecture. For the proposed architecture in this paper, MLC-GCN with 12 ATSFEs has an accuracy of over 80\% and with 24 ATSFEs reaches 82.26\% on ADNI, higher almost 10\% than GCN. The performance of MLC-GCN shows an upward trend with the increasing ATSFEs indicating effectiveness of the deep features when the data become complex for the multi-classification task.

The result of OASIS-3 is similar with ADNI. The strategy of multi-level feature extraction to construct connectomes intends to simultaneously expand the depth of information exploration and the division of feature levels in the model. Table \ref{tab.3} indicates the model performance improves as the levels of STFE upsend, showing effectiveness of the sparse advanced features. MLC-GCN with three different stacks  MLC\_GCN$_6$,  MLC\_GCN$_{12}$ and  MLC\_GCN$_{24}$ get the best performance among all the compared methods. It should be noted that  the results on OASIS-3 are nearly overall higher than those on ADNI as the labels number on OASIS-3 is 3 in our experiments.

\begin{figure}[!t]
	\centering
	\includegraphics[width=1.0\columnwidth]{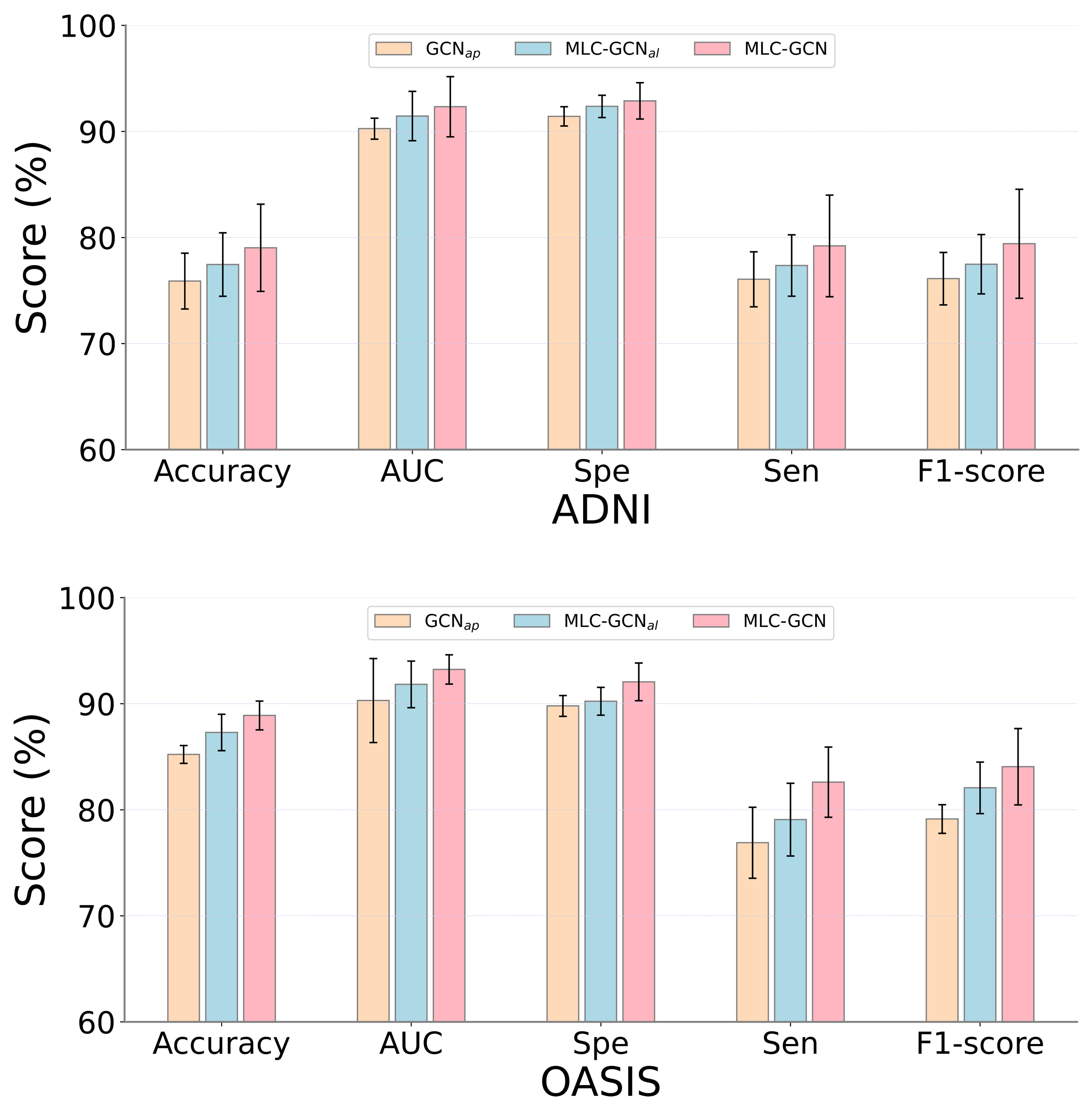}
	\caption{Ablation study of MLC-GCN with 5-fold cross validation by 5 evaluation measurements. }
	\label{fig.3}
\end{figure}

\begin{figure}[!t]
	\centering
	\includegraphics[width=1.0\columnwidth]{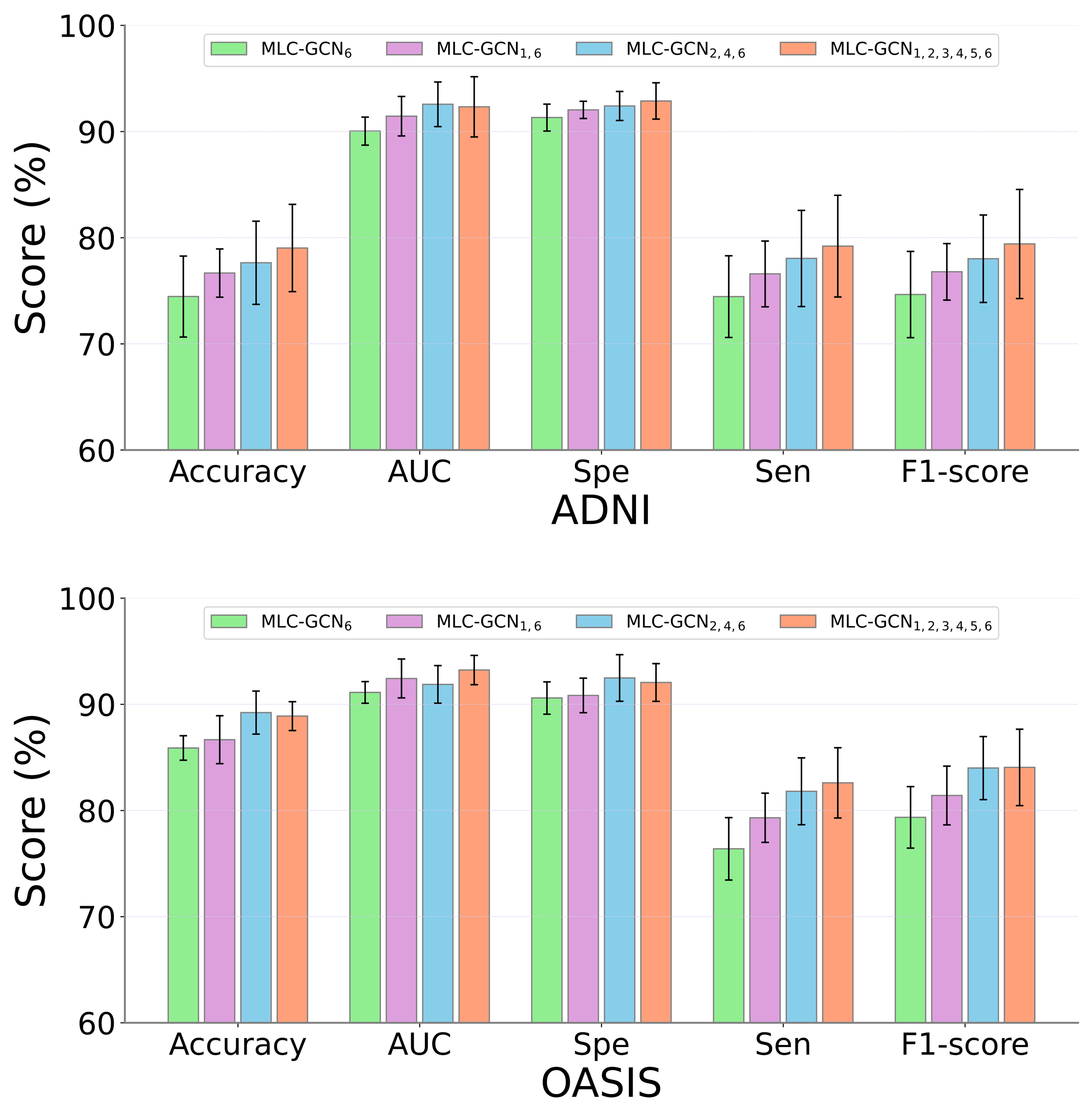}
	\caption{Ablation study of MLC\_GCN$_6$ of selected feature levels with 5-fold cross validation. The subscript indicates the index of the feature level with "1" the initial STFE level and "6" the deepest one.}
	\label{fig.4}
\end{figure}

\begin{figure}[!t]
	\centering
	\includegraphics[width=1.0\columnwidth]{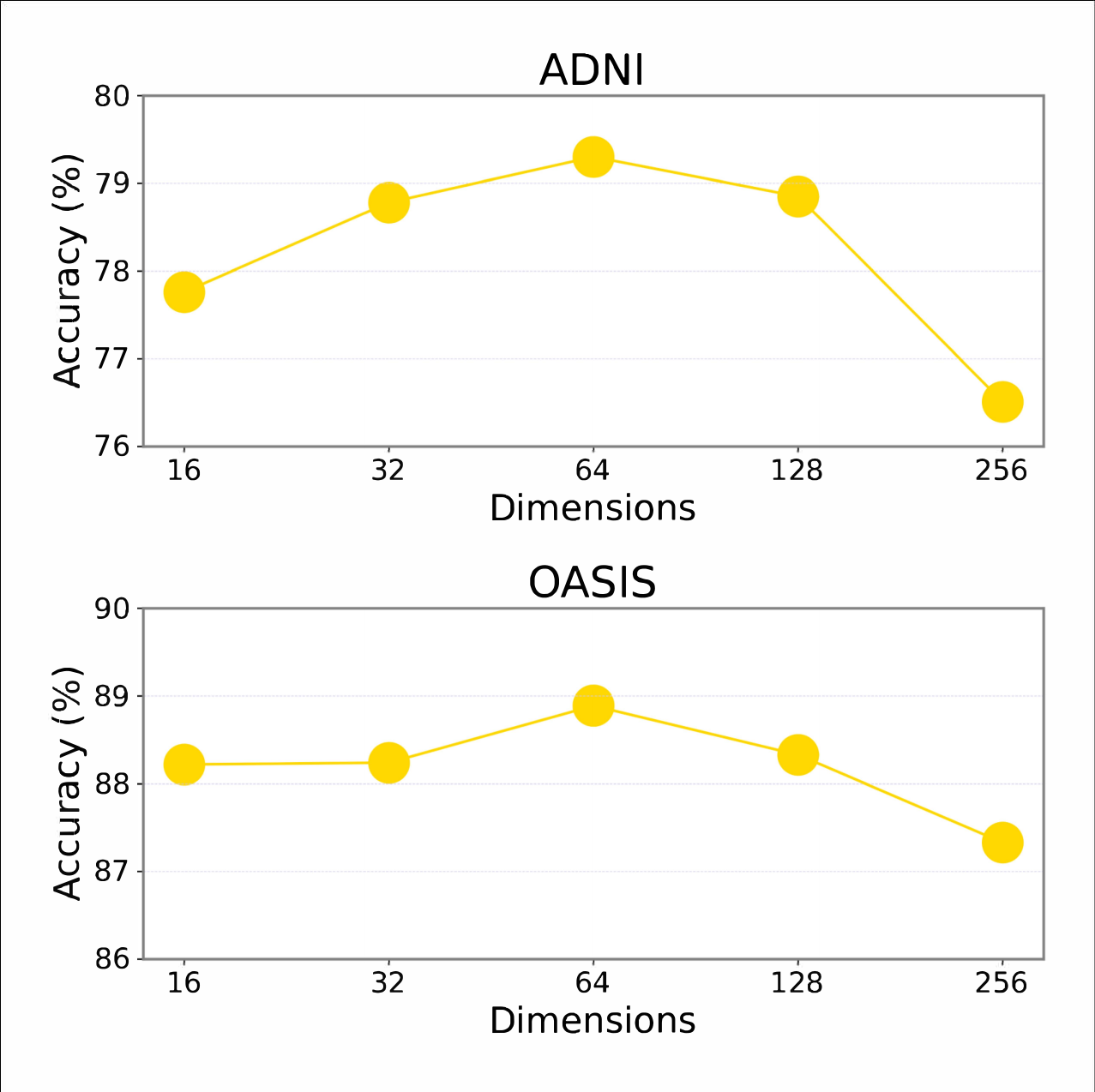}
	\caption{Ablation study of different numbers of embedding length $l$ with 5-fold cross validation.}
	\label{fig.5}
\end{figure}

\begin{figure*}[!t]
	\centering
	\includegraphics[width=1.5\columnwidth]{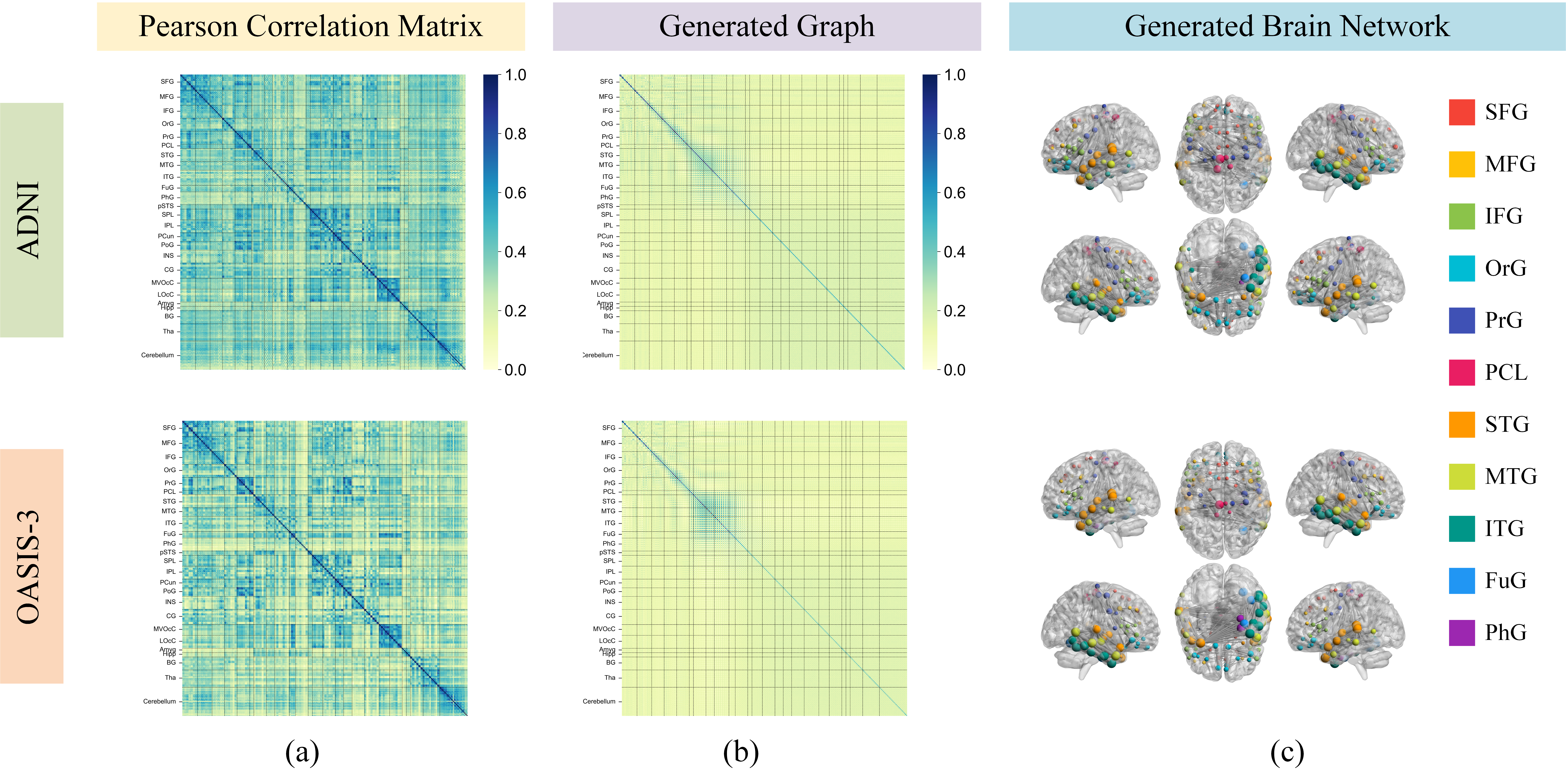}
	\caption{Visualizations of FC matrix with Pearson correlation of handcrafted connectome and one of the generated graphs on the normalized heatmap as well as the connectomes. (a) Pearson correlation matrix, (b) generated graph, (c) generated brain network with colors indicating different ROIs.}
	\label{fig.6}
\end{figure*}

\begin{figure*}[ht]
	\centering
	\includegraphics[width=1.5\columnwidth]{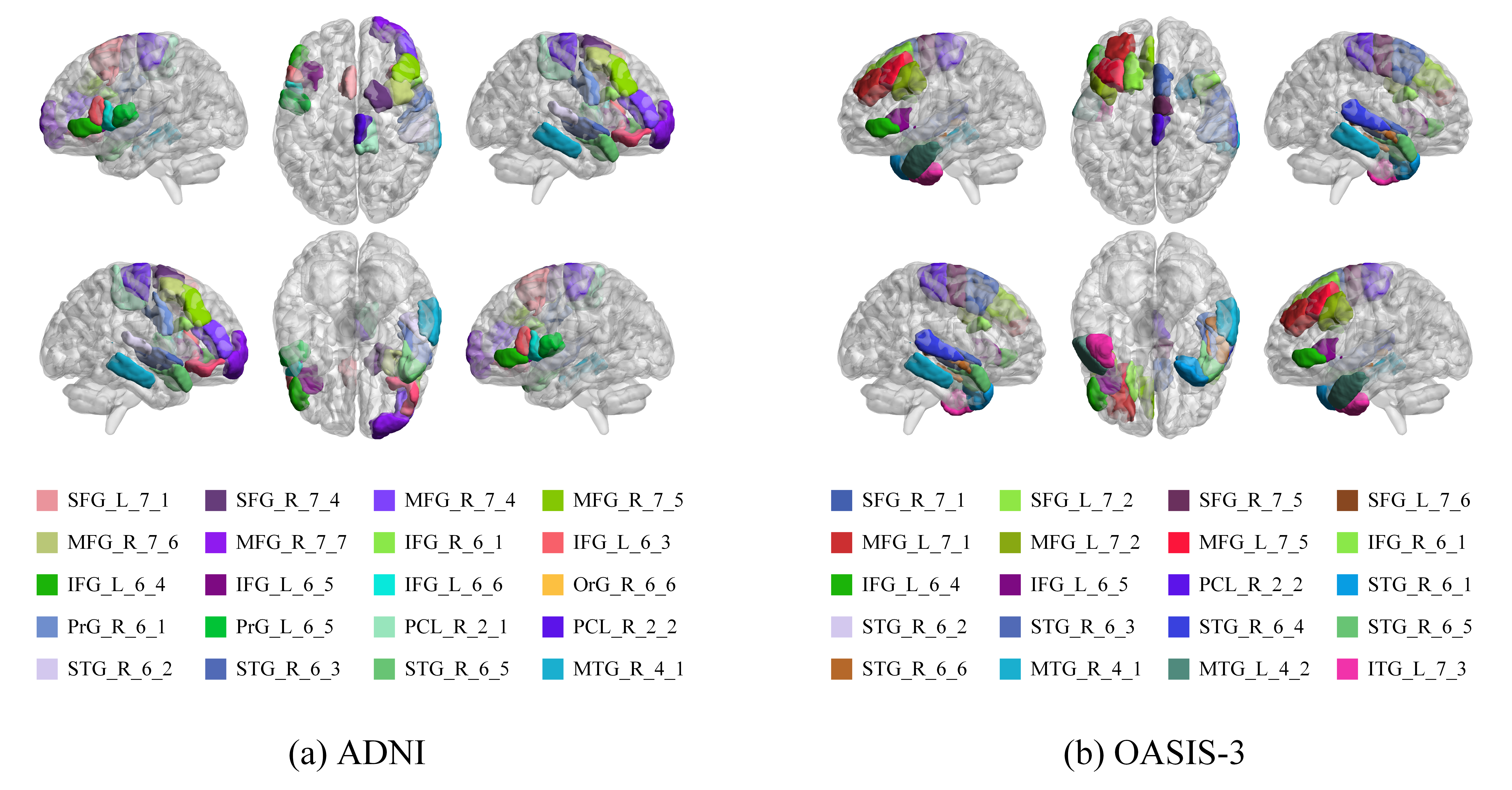}
	\caption{Top 20 most important brain regions associated with AD on (a) ADNI and (b) OASIS-3 in the generated connectomes.}
	\label{fig.7}
\end{figure*}

\subsection{Ablation Experiments}
\label{sec:3.5}
Table \ref{tab.4} shows the ablation experiments to validate the designed components for the graph generator in MLC-GCN. The performance of MLC-GCN decreases when removing TFE or SFE, which implies their strong temporal and spatial feature extraction separately in the graph generator. In addition, the performance reduces more by removing SFE than by TFE, which may indicate that the temporal features are less important than the interregional correlation. These results prove the  appropriateness of taking the spatial connectivity into account during the graph generation proposed in the paper. The regularization ${L_{group}}$ also improves the performance of the model to a certain extent by controlling the generator to cluster the graphs according to the labels. MLC-GCN with TFE, SFE and ${L_{group}}$ achieves the highest performance on all metrics indicating that each component contributes to improve of the architecture.

Fig. \ref{fig.3} shows a comparison of the results of MLC-GCN including 6 ATSFEs with the other two models to verify the usage of multi-layer GCN encoders. $\text{GCN}_{ap}$ includes 7 identical layers as all edges of the graphs in different layers are composed of Pearson correlation matrices, and $\text{MLC-GCN}_{al}$ adopts the same architecture with MLC-GCN, but all graphs are generated from the output of the last STFE. MLC-GCN is higher than $\text{GCN}_{ap}$ across all metrics on ADNI and OASIS-3 showing that the generated graphs capture even better feature than that of Pearson matrices. Moreover, the difference between MLC-GCN and $\text{MLC-GCN}_{al}$ on performance also indicates different levels of features more effectively represent the fMRI data than the single-layer features.

Fig \ref{fig.4} presents the ablation results on the different number of levels in graph generators of MLC-GCN with 6 ATSFEs. The results on two datasets demonstrate that the concatenating brain graphs at different levels can effectively improve the performance of classification. MLC-GCN with 6 graphs from each STFE far outperforms $\text{MLC-GCN}_{6}$ with a single brain graph generated from the 6th level, and still is higher than the architectures including a part of the levels. $\text{MLC-GCN}_{2,4,6}$ involving the 2nd, 4th and 6th levels gets a lightly higher score on $Accuracy$ and $Spe$ than MLC-GCN on OASIS-3, revealing the certain inter level gap may also be beneficial to feature selection and the more relaxed architecture may alleviate overfitting. These experiments imply the multi-level graphs proposed in this paper capture more the hierarchical information of the brain in fMRI than the single or a few ones.

Fig \ref{fig.5} displays the results of different numbers of embedding length $l$ for MLC-GCN. The accuracy on two datasets increases with the increasing numbers of embedding length from 16 to 64, and then reversely decrease. The performance of the model reaches its peak when the embedding length is selected between 32 and 128. The reason can be that a smaller length can lead to information loss during the embedding process, making it unable to contain enough useful information and a larger length, especially longer than the original time series, contains too much loose information, which hinders the model's ability to integrate feature information. In this observation, we set the numbers of embedding length to 64 in all other experiments.

In fact, MLC-GCN has some universality, and we only adopt the standard GCN as the encoder, which can be further improved by considering more powerful GNN encoders.

\subsection{Model visualization}

To check what the MLC-GCN learned under the regularization of both cross entropy loss and intra-class graph dissimilarity, we visualized the mean learned connectome graphs of the multi-graph generator module after MLC-GCN training.  Fig. \ref{fig.6} (a) and Fig. \ref{fig.6} (b) display the mean correlation coefficient matrix of the non-STFE processed fMRI time series ($G_)$) and the mean of all generated graphs of all subjects ($G_1$ to $G_K$).  The generated graph shows higher sparsity than the graph of the non-temporal feature processed (non-STFE processed) time series with high connectivity mainly located in prefrontal and temporal cortex (Fig. \ref{fig.6} (c)). Fig. \ref{fig.6} (c) shows the graph in the 3D brain space. Network visualization was performed using the BrainNet Viewer (http://www.nitrc.org/projects/bnv/) \cite{xia2013brainnet}.  The graph is generated by the top 1\% most connected edges of the generated connectomes shown in Fig. \ref{fig.6} (b). The size of node indicates the number of connected edges after thresholding and the color indicates different brain ROIs.

Fig. \ref{fig.7} visualizes the top 20 most important brain regions associated with AD in the MLC-GCN models built for ADNI and OASIS-3 separately. The importance of ROIs is quantified by summing the edge weights of each node in the average generated graph. The two independently trained MLC-GCN models show over-lapped top 20 nodes in superior frontal gyrus (SFG), middle frontal gyrus (MFG), inferior frontal gyrus (IFG), paracentral lobule (PCL), superior temporal gyrus (STG) and middle temporal gyrus (MTG).

\section{Conclusion}
\label{sec:5}
In this paper, we proposed a stack of spatio-temporal feature extraction and graph generation based clinical outcome prediction model. Through the intra-class graph dissimilarity regularization, the multi-level STFEs and GCNs are trained to learn different sparse brain graphs at different scales. Through independent cohort validations, MLC-GCN shows better performance for differentiating MCI, AD, and normal aging than state-of-art GCN and rsfMRI based AD classifiers. The proposed MLC-GCN also showed high explainability in terms of learning clinically reasonable connectome node and connectivity features from two independent datasets. While we only tested MLC-GCN on AD, the basic rsfMRI-based multi-level learned GCN based outcome prediction strategy is valid for other diseases or clinical outcomes.

\bibliographystyle{IEEEtran}
\bibliography{ref}

\end{document}